# AUTONOMOUS RENDEZVOUS WITH NON-COOPERATIVE TARGET OBJECTS WITH SWARM CHASERS AND OBSERVERS

Trupti Mahendrakar[*], Steven Holmberg[†], Andrew Ekblad[‡], Emma Conti[§], Ryan T. White[**], Markus Wilde[††], and Isaac Silver[‡‡]

Space debris is on the rise due to the increasing demand for spacecraft for communication, navigation, and other applications. The Space Surveillance Network (SSN) tracks over 27,000 large pieces of debris and estimates the number of small, un-trackable fragments at over 1,00,000. To control the growth of debris, the formation of further debris must be reduced. Some solutions include deorbiting larger non-cooperative resident space objects (RSOs) or servicing satellites in orbit. Both require rendezvous with RSOs, and the scale of the problem calls for autonomous missions. This paper introduces the Multipurpose Autonomous Rendezvous Vision-Integrated Navigation system (MARVIN) developed and tested at the ORION Facility at Florida Institution of Technology. MARVIN consists of two sub-systems: a machine vision-aided navigation system and an artificial potential field (APF) guidance algorithm which work together to command a swarm of chasers to safely rendezvous with the RSO. We present the MARVIN architecture and hardware-in-the-loop experiments demonstrating autonomous, collaborative swarm satellite operations successfully guiding three drones to rendezvous with a physical mockup of a non-cooperative satellite in motion.

## INTRODUCTION

With the increasing amount of on-orbit near-earth spacecraft, there have been increasing amounts of space debris. There are currently over 27,000 space debris objects being tracked by the Department of Defense's global Space Surveillance Network (SSN) out of which, some are old decommissioned non-cooperative spacecraft and others vary from fragments of spacecraft or rocket boosters to even chips of paint traveling at very high speeds [1]. Collision with any of these non-cooperative Resident Space Objects (RSO) could cause catastrophic damage to a mission.

The most efficient way to reduce the amount of space debris is by preventing the formation of future space debris. One effective approach for prevention is performing on-orbit servicing (OOS) to extend the life of a retired spacecraft as recently demonstrated by Northrup Grumman's Mission Extension Vehicles [2, 3]. Doing so minimizes the additional need to launch existing functionality and further formation of debris. Another effective approach is to deorbit existing uncorrelated spacecraft that are non-functional, whose orbital positions could be estimated from the Space Surveillance Network (SSN) data, but the structure, functionality, and capabilities are unknown.

---

[*] PhD Candidate, Aerospace Engineering, Florida Institute of Technology, 150 W University Blvd, Melbourne, FL, US.
[†] Undergraduate Student, Aerospace Engineering, Florida Institute of Technology, 150 W University Blvd, Melbourne, FL, US.
[‡] Masters Student, Electrical Engineering, Florida Institute of Technology, 150 W University Blvd, Melbourne, FL, US.
[§] Undergraduate Student, Aerospace Engineering, Florida Institute of Technology, 150 W University Blvd, Melbourne, FL, US.
[**] Assistant Professor, Mathematical Sciences, Florida Institute of Technology, 150 W University Blvd, Melbourne, FL, US.
[††] Associate Professor, Aerospace Engineering, Florida Institute of Technology, 150 W University Blvd, Melbourne, FL, US.
[‡‡] Chief Scientist, Energy Management Aerospace LLC, 2000 General Aviation Drive, Hangar 10I, Melbourne, FL, US.

Some prior OOS/inspection missions that are similar to the research discussed in this paper include ELSA-d [4], XSS-10/11 [5, 6], ETS-VII [7], and ANGELS [8]. The typical approach for these missions has been to use a single chaser spacecraft to rendezvous and dock with a familiar RSO whose geometry and functionality are fully known. However, active space debris removal (ADR) and OOS as studied in this work involves a large non-cooperative spacecraft whose structure and functionality may be unknown in advance. To complicate matters, the spacecraft could have a significant tumbling rate and unknown safe capture interfaces, increasing the complexity of safe approach trajectories.

Additionally, using a single chaser to capture and stabilize such RSO would create large forces and moments on the capture interfaces. As an alternative, swarm satellites that can autonomously identify capture interfaces on the RSO and autonomously navigate to dock with the RSO to minimize the tumbling rates would significantly lower the resultant forces and moments associated with docking. This concept is depicted in Figure 1.

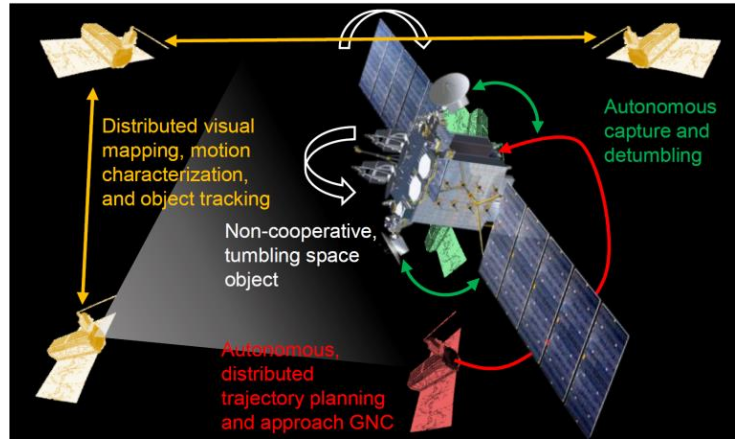

**Figure 1: Autonomous Docking with Swarm Satellites**

Based on this concept, our previous research demonstrated a real-time, on-board machine vision system could be used to successfully localize solar panels and bodies of a target satellite in a camera feed [9], estimate a sparse 3D point cloud labeling solar panels as repulsive points and bodies as attractive points for an artificial potential field (APF) guidance algorithm, and simulate a single chaser to safely dock with an RSO with Simulink [10]. Building on our previous work, this paper presents the Multipurpose Autonomous Rendezvous Vision-Integrated Navigation (MARVIN) integrated hardware/software system that performs navigation and guidance to command swarms of drones to dock with an RSO using low-power, spaceflight-equivalent computers. This paper presents the results of end-to-end autonomous hardware-in-the loop testing using the MARVIN system in the ORION Facility at Florida Institution. This work demonstrates the MARVIN system can successfully navigate and guide three drones representing small chaser satellites to dock with a non-cooperative RSO.

**RELATED WORK**

This work draws inspiration from developments in both autonomous navigation and deep learning-based computer vision, both fast-evolving fields.

### Artificial Potential Field Guidance

APF path planning is popularly used in the field of robotics, originally developed to guide robots around static objects [11] and for unmanned air vehicle operations [12]. The methods have been gaining interest in the field of swarm satellite operations recently. Several articles that have leveraged the APF methods for swarm satellite operations are: [13] used APF for swarm satellites with sliding mode controllers, [14, 15] used APF for formation flying around flexible spacecraft and [16, 17, 18] used APF to perform close proximity operations around non-rotating targets.

Of particular interest, APF guidance laws have been shown to enable a swarm of small chaser spacecraft to approach and capture a non-cooperative RSO collaboratively and safely. In prior work [19], a virtual potential field was constructed by locating virtual repulsive field sources at collision hazards and other keep-out areas and locating attractive field sources at capture locations or inspection positions, and chasers simply follow potential gradients to the target. In this article, four chasers approached a rotating target object from arbitrary starting positions and were shown in simulations to safely capture the target with realistic limitations on $\Delta v$, required thrust, and velocity on capture contact.

### Deep Learning-based Machine Vision

The fast-paced advances in deep learning algorithms and computing hardware over the past 10 years have made highly accurate computer vision algorithms available on sufficiently low compute and power budgets for on-board applications in space. Specifically, deep convolutional neural networks [20] accelerated by massive parallelization of tensor arithmetic on graphics processing units (GPUs) [21] led to tremendous accuracy gains in vision benchmarks in image classification [22], object detection [23], and image segmentation among other tasks. The algorithms at times even outperforming humans.

While these feats were initially restricted to conventional computing hardware, the joint development of GPU-accelerated edge computers and computer vision algorithms designed or optimized for them [24] [25] have opened the doors to edge deployment applications (e.g. in space, robots, autonomous vehicles, sensor networks). These developments allow the present work to develop and test a system to perform object detection on low-compute, low-energy edge hardware approximating the capabilities of spaceflight computers.

Important for this work is the object detection task in computer vision. Object detection requires identifying and localizing objects in a camera frame—that is, to predict a pixel-level bounding box tightly surrounding each object from a pre-specified set of classes and classify each object. State-of-the-art object detection algorithms fall into three categories: vision transformers, multi-stage object detectors, and single-stage object detectors.

Vision transformers [26] are the most accurate but have computing requirements that generally cannot be satisfied by current edge hardware. Multi-stage detectors work by breaking the object detection task into multiple sub-problems. For example, multi-stage object detector Faster R-CNN [27] has two networks: one that proposes bounding boxes that might surround objects and one to determine if the box surrounds an object and classify it. In contrast, single-stage object detectors perform object detection with a single network predicting both bounding boxes and object classes simultaneously. The You Only Look Once (YOLO) algorithm [25] and its successors are the best performing single-stage detectors.

Both single-stage and multi-stage object detectors can run on the edge GPUs chosen for this work. However, prior research [28] demonstrated Faster R-CNN could only achieve framerates too low for real-time use in rendezvous operations. However, the YOLOv5 [29] single-stage object detector was shown to be effective for real-time satellite feature recognition (including solar panels, bodies, antennas, and thrusters) in the lab under a variety lighting and motion conditions for the observer and target RSO [9], even with significant variation of solar panels [30] with much better framerates.

**Fusing YOLOv5 with APF Guidance**

Our prior work [10] demonstrated YOLOv5 detections with predicted class labels localized in 3D with a stereographic camera in the lab could feed an APF flightpath planning algorithm for successful rendezvous with a rotating target with a single chaser simulated with MATLAB/Simulink. The present work extends these ideas and presents the complete MARVIN hardware-software system, which is shown to perform end-to-end hardware-in-the-loop experiments where real-life chaser drones guided by an APF guidance algorithm. In the experiments presented below, this approach is shown to safely avoid fragile solar panels and dock with the body of a target RSO.

**THE MARVIN SYSTEM**

MARVIN system includes two integrated subsystems, machine vision and APF. These subsystems work together to autonomously navigate and guide chaser drones to an unfamiliar non-cooperative RSO. The system-level architecture of MARVIN is shown in Figure 2. Each subsystem is equipped with off-the-shelf hardware. The machine vision subsystem has a Raspberry Pi 4B (8 GB RAM), an Intel RealSense D435i camera (stereographic camera), and an Intel Neural Compute Stick (NCS) 2. The APF subsystem is equipped with another Raspberry Pi 4B (8 GB).

The requirements that governed the development of MARVIN:

1. The machine vision subsystem must detect and localize all visible solar panels and the body of the target RSO in a camera feed with no prior knowledge of the RSO.
2. The machine vision subsystem must localize all RSO components in 3D using its stereographic camera.
3. The machine vision subsystem must pass the 3D positions of visible components to the APF subsystem via ROS 2.
4. The APF subsystem must accept 3D positions of each component from the machine vision subsystem via ROS 2.
5. The APF subsystem must receive current positions of the chaser drones via UDP.
6. The APF subsystem must generate safe approach trajectories and corresponding chaser accelerations to the RSO body within the limits of a collision-avoidance radius.
7. The APF subsystem must communicate guidance accelerations to the chasers in real time.

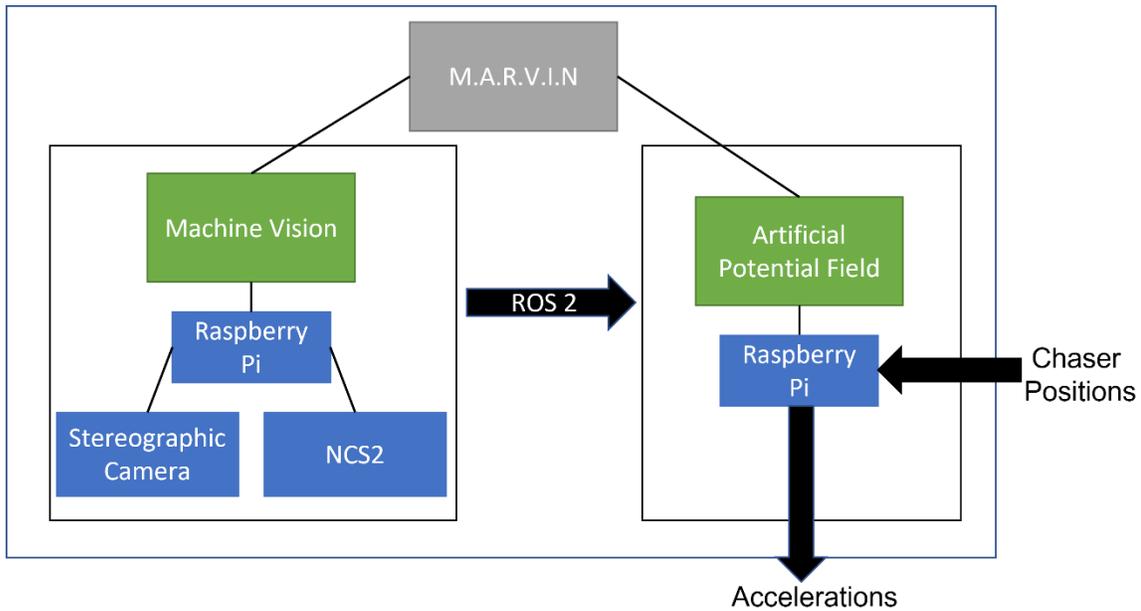

**Figure 2: MARVIN System Architecture**

## MACHINE VISION-BASED NAVIGATION

As mentioned in the introduction, the machine vision algorithm developed for this research identifies solar panels and the body of the RSO. (Note that it can be used for antennas and thrusters as well, but these components are not used in the current article.) The algorithm is trained on a set of satellite images gathered in the public domain gathered for [9] and amended for [28], *not* including any images of the satellite mockup we use in the lab. Rather, all experiments run a trained vision algorithm that is operating with an RSO the model has never seen, hence demonstrating our approach is agnostic to the target being observed. Therefore, our solution can be implemented with any unknown target RSO with such components.

### Machine Vision Subsystem

The subsystem consists of a Raspberry Pi 4B (8 GB RAM), Intel Neural Compute Stick 2 (NCS2), and an Intel D435i RealSense stereographic depth camera. This subsystem provides a constrained computational environment with low power draw similar to spaceflight computers. Since the goal of this work is to enable close proximity operations, the RealSense camera's approximately 5-meter effective range is reasonable, although a larger camera or an additional observer could triangulate the distances with a further range.

The RealSense camera provides the Pi with a video feed. Each frame of the video is a color image with estimated depth at each pixel (i.e., RGB-D data). The YOLOv5 algorithm pretrained on satellite imagery in [30] runs real-time inference using the RGB channels of the video frame on the Pi and NCS2, meaning it classifies each component and localizes each with a bounding box. The Pi then reads in these bounding boxes (as shown in Figure 3) and depth channel of the video frame. From this, the 3D positions of 5 points in each bounding box at the centroid and near each corner of the are computed. The 5 points are shown in each bounding box in Figure 3, and correspond to points labeled P1, P2, P3, P4, and P5 in Figure 4. For each bounding box, the 3D positions of these 5 points and predicted component type is broadcasted over the ROS2 [31] platform.

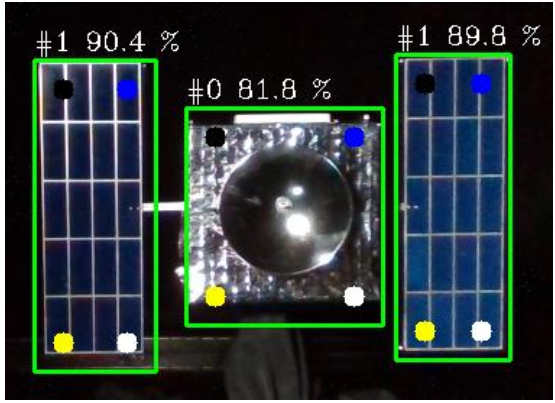 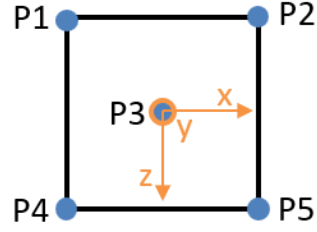

**Figure 3: Picture of machine vision detection**    **Figure 4: Bounding Box Points**

## ARTIFICIAL POTENTIAL FIELD GUIDANCE

The APF subsystem consists of a Raspberry Pi 4B (8 GB RAM) meant to simulate the computational and power constraints of spaceflight computers.

The APF subsystem receives two sets of inputs: (1) 3D bounding box information and (2) live chaser position and velocity data. The bounding box information from the machine vision subsystem includes 3D points with class labels (solar panel or body). To receive information, the APF subsystem subscribes to the machine vision broadcast to receives the 3D bounding box points and labels and similarly receives live chaser position and velocity data. The live chaser data comes from outside the MARVIN system. This data feed may come directly from the chasers or from a centralized tracker, although the experiments below have MARVIN subscribe to live chaser data from an OptiTrack tracker ROS2 broadcast.

Based on the two inputs feeds, the APF algorithm updates its repulsive points (solar panels) and attractive points (bodies) in real-time and safely guides the chasers to selected docking positions on a target satellite by computing chaser accelerations and communicating them to the chasers.

### APF Algorithm

As discussed in [10], the APF algorithm utilizes the potential field acceleration function as shown in Equation 1, formulated in [19]. $\gamma$ is the chaser control acceleration due to the known position of the target calculated by super positioning the potential field contributions of $k$ repulsive potential field nodes and $l$ attractive potential field nodes. $r_c, \dot{r}_c$ are the position and velocity terms of each chaser spacecraft relative to the centroid of the target spacecraft and $r_{target}$ is the position array of the target spacecraft nodes. The relative positions for the chaser with respect to each repulsive or attractive field node is represented by $\boldsymbol{\rho}_i$ and $\boldsymbol{\rho}_j$.

The acceleration contribution of each repulsive node is the product of its repulsive field gain, $\mu_R$, and the difference between the out-of-bounds radius, $r_d$, and the distance between the chaser and the node (R-switch), $\rho_i$. As a result, if the distance of the chaser exceeds the allowable distance from the target, the $(r_d - \rho_i)$ term flips to negative. The contribution of an attractive node is the product of the attractive field gain, $\mu_A$, and the distance between the chaser and the node, $\rho_j$. Attractive node acceleration is consistently attractive despite the location within the potential field. Additionally, both the repulsive and attractive nodes are influenced by a velocity dampening term $c(\dot{\boldsymbol{r}}_c \cdot \boldsymbol{\rho}_i)$ and $c(\dot{\boldsymbol{r}}_c \cdot \boldsymbol{\rho}_j)$, where $\dot{\boldsymbol{r}}_c$ is the chaser velocity vector, $\boldsymbol{\rho}$ is the chaser position vector relative to each node, and *c* is the field gain scalar. The scalar acceleration contribution is finally multiplied by the relative position unit vector $\hat{\boldsymbol{\rho}}_i$ or $\hat{\boldsymbol{\rho}}_j$ for direction.

$$\boldsymbol{\gamma}(r_C, \dot{r}_C, r_{target}) = -\sum_{i=1}^{k}[\mu_{R,i}(r_d - \rho_i) + c(\dot{r}_c \cdot \boldsymbol{\rho}_i)]\hat{\boldsymbol{\rho}}_i - \sum_{j=1}^{l}[\mu_{A,j}\rho_j + c(\dot{r}_c \cdot \boldsymbol{\rho}_j)]\hat{\boldsymbol{\rho}}_j + \boldsymbol{\gamma}_{cc} \qquad (1)$$

The final acceleration contribution in Equation 2 is due to chaser-chaser interactions, $\boldsymbol{\gamma}_{cc}$, where *m* is the total number of chasers, $\mu_c$ is the chaser-chaser field gain scalar, $\rho_c$ is the distance between chasers, and *X* is a conditional variable that determines if a chaser is within the collision avoidance radius of another chaser. The function only evaluates over m-1 because the chaser is repelled by other chasers, excluding itself. $\mu_c$ is always a negative value to produce a repulsive acceleration between chasers. The repulsion also scales with distance, greatest when chasers are closer together. Finally, the conditional *X* exists to only add chaser-chaser repulsion when a chaser is within a preset distance of another chaser; if a chaser is too far away, *X* is set to zero. *X* is meant to prevent chasers from influencing each other when on opposing sides of a target spacecraft.

$$\boldsymbol{\gamma}_{cc} = -\sum_{n=1}^{m-1}\frac{\mu_c}{e^{\rho_c}}X \qquad (2)$$

The field acceleration vector $\boldsymbol{\gamma}$ is fed into Hill's equations represented in Equation 3 to calculate the final swarm chaser acceleration vectors which are commanded to each chaser to execute the mission.

$$\ddot{r}_C = \begin{bmatrix} 2\omega\dot{z}_C + \dot{\omega}z_C \\ -\omega^2 y_C \\ 3\omega^2 z_C - 2\omega\dot{x}_C - \dot{\omega}x_C \end{bmatrix} + \boldsymbol{\gamma} \qquad (3)$$

The target RSOs orbital rate is represented by $\omega$ and, $x_c, y_c, z_c$ are the positions of the chasers respect to the RSO and, $\dot{x}_c, \dot{y}_c, \dot{z}_c$ are the relative velocities of the chasers with respect to the RSO in the LVLH frame (referred as the APF frame in the paper). In the LVLH frame, x is along the velocity direction of the RSO, *z* is in the nadir direction pointing to the earth and, *y* is out of plane, cross product of *x* and *z* directions.

**EXPERIMENTAL SETUP**

The experiments discussed in this paper are conducted at the ORION facility at the Florida Institute of Technology. The ORION facility is equipped with hardware-in-the-loop formation flight simulator and a docking simulator [31]. The past work which formed the foundation of this research was also tested at the ORION facility.

**MARVIN Experimental Specifications**

To produce an effective and responsive guidance system, the cartesian coordinates provided by machine vision are converted into nodes that construct an artificial potential field (APF). The APF algorithm converts different coordinate frames from machine vision and OptiTrack into its own reference frame. The coordinate frames for the experiment are shown in Figure 5. The APF coordinate system is centered with the target. The frame was kept constant whether the target satellite was tumbling or holding a steady position.

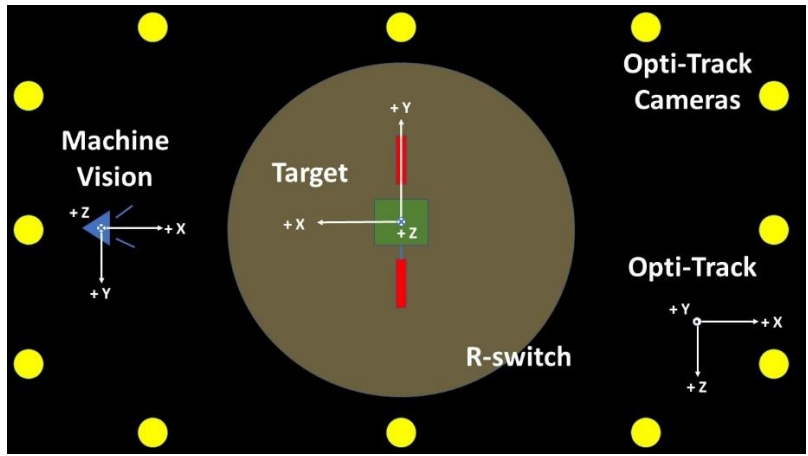

**Figure 5: Coordinate System**

To construct the field, the APF algorithm assigns each node an attractive index potential. Nodes located on undesirable locations, such as solar panels, are assigned negative values, whereas, desirable positions, such as potential docking sites, are assigned positive values. Additionally, repulsive nodes are assigned to the live positions of active chasers to prevent collisions. Chaser-Chaser repulsion, however, only contributes to the potential field acting on a chaser when their collision avoidance radius is breached. The field gain values for each of the nodes were initially based on simulations however when the scale of the experiment was changed and errors were introduced, the simulation values did not perform well. As a result, the existing field gain values were experimentally derived. The relative position of each node to a chaser and their corresponding attractive index are weighed to build an acceleration vector. The summation of all the acceleration contributions from each of the nodes built a path acceleration vector for a chaser to follow, see Equation 1 and Equation 2.

The resultant acceleration is continuously updated and exported to chasers. As a result of the rapid refresh rate of the APF algorithm, chasers were able to react quickly to dynamic changes such as avoiding chaser collisions and docking with a moving target. Moreover, the resultant potential field had conditional components. To prevent chasers from flying too far away from the target, an "out-of-bounds" region, R-Switch, was put into the system so that in the event a chaser was pushed too far, all negative nodes would flip positive to pull the chaser back into the range, see Equation 1. Also, to prevent chaser-target collisions, a velocity dampening acceleration exists to slow down a chaser when making its final approach, see Equation 1.

For this experiment, both the subsystems are integrated by a network card over which all the communications are passed. Figure 6 shows an overview of all the systems working together in the experiment. A notable difference between Figure 6 and Figure 2 is that the APF algorithm for this setup passes future position and velocity data to the drones instead of acceleration due to software limitations of the drones.

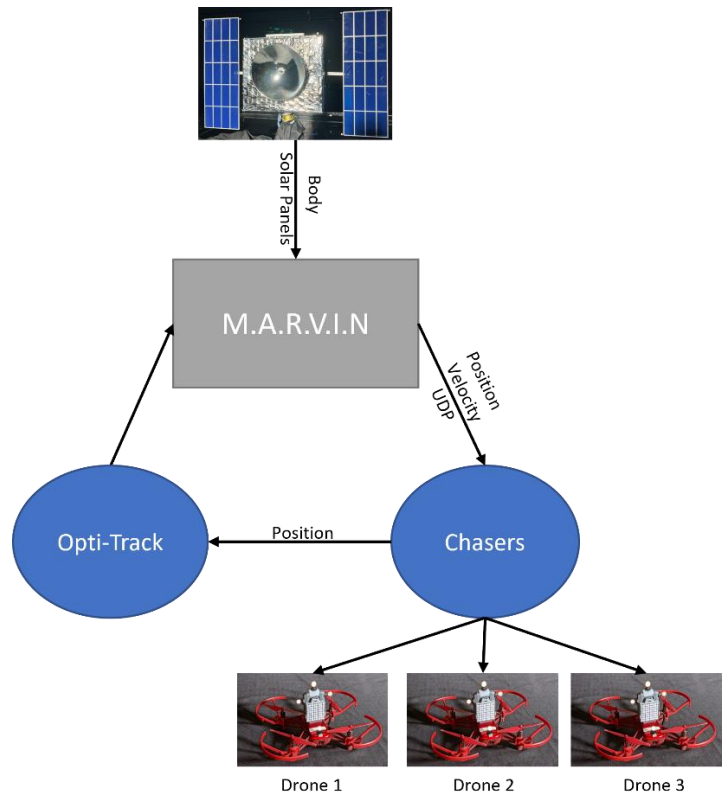

**Figure 6: Experiment Setup - Flow Diagram**

**OptiTrack**

While the machine vision subsystem streams the live position of the target satellite to APF, OptiTrack cameras, coupled with Motive software, provides the algorithm with the live positions of the chasers. A separate algorithm retrieves the rigid body data from OptiTrack and publishes the data over a LAN using ROS2.

**Chasers**

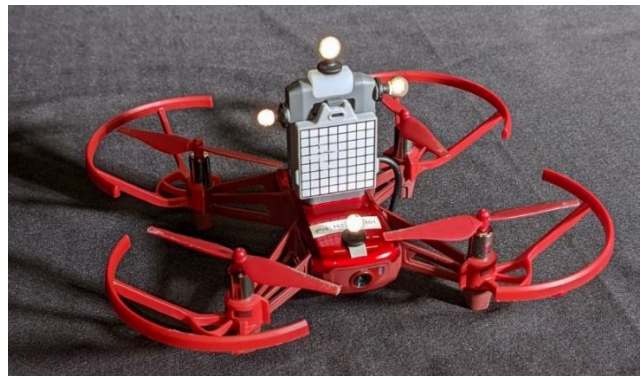

**Figure 7: DJI Robomaster Tello Talent Drone**

DJI Robomaster Tello Talent drones shown in Figure 7 were utilized in the laboratory setting to simulate chaser satellites [32]. The Tello drones are 18 cm long and wide and 9 cm tall. The Tello drones were chosen for their programmability and their swarming feature: being able to command multiple drones from one computer. Swarming is possible because each Tello drone has been setup to automatically connect to a LAN where they are auto assigned a specific IP address based on their MAC address. The APF algorithm is then able to export individual drone commands by sending them over UDP to each drones' unique IP address. The drones perform translation movements in their own cartesian coordinate frame when given a position (cm) and speed (cm/s). The drones are limited in the distance they can move in any given command: X/Y/Z movements must be within –500 and 500 cm and cannot be between –20 and 20 cm, and speeds must be within 10 and 100 cm/s. If a command out of the provided bounds is received, an "out of bounds" error will occur, and the drone will ignore the command. Also, if another command is sent before the drone has completed the previous command, it will ignore the command and will accept the next command sent after its movement is completed. Each one does have a maximum flight time of seven minutes and has an infrared camera and an IMU built in to stop and stabilize themselves after performing a movement.

**Integration**

Using the existing location of the known nodes, two more attractive nodes are added to the body of the target on the unseen, adjacent face. The target image is rebuilt in the algorithm by calculating the target's centroid and setting up the nodes around the centroid. The target node positions are then inflated using a safety scale of 1.75 to prevent collisions from drones overshooting and to account for the space between the nodes and the bounding box generated by the machine vision subsystem. Since the chasers require position and speed to react, the acceleration is integrated into commands that the drones can understand; the drones also have their own coordinate system with Y and Z opposite to APF. To increase accuracy, the drone position commands were rounded down to the minimum movement (-20 or 20 cm) and the speed was set to its maximum (100cm/s). By minimizing the drone's movement and the drone's maximizing speed, it allowed the algorithm to refresh drone commands quicker and aid in preventing chaser-chaser collisions. Higher chaser-chaser repulsion accuracy also greatly reduced errors from drones blowing each other off-course and reduced error due to integration.

A checker evaluates the position of each of the chasers and the attractive nodes to determine if a chaser is within range of docking. Once a chaser is within the predetermined range, it freezes all chaser movement to complete docking and avoid collisions or interference from chaser-chaser repulsion. After the chaser remains in range for a set amount of time, the chaser is removed from the experiment and lands itself. After all the drones have completed the simulation, the function ends by graphing the detected positions, directly from OptiTrack, to show the path taken by each chaser and the last position of the target. Finally, Hill acceleration was excluded during these tests due to small scale of the experiment and because the experiment was conducted at sea-level.

**CHASER SWARM EXPERIMENTS**

The APF coefficients used for the hardware-in-the-loop experiments are tabulated in Table 1 and the experimental set-up at the ORION facility is shown in Figure 8.

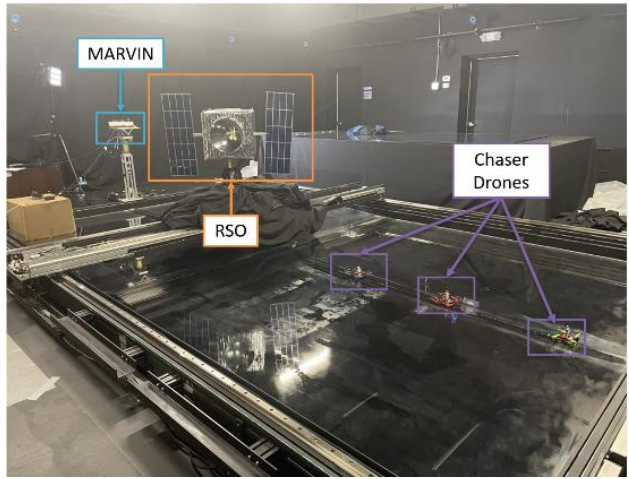

**Figure 8: Major pieces of the ORION experiment setup**

**Table 1: APF Coefficients**

| Variable | | Equation | Experimental Weight |
|---|---|---|---|
| Attractive Field Gain | $\mu_A$ | 1 | 0.1 |
| Repulsive Field Gain | $\mu_R$ | 1 | -0.015 |
| R-Switch (m) | $r_d$ | 1 | 2.0 |
| Velocity Dampening Scalar | c | 1 | 0.08 |
| Chaser-Chaser Field Gain | $\mu$ | 2 | -2.5 |

The values were experimentally derived. An R-Switch value of 2 meters was chosen due to the size of the lab setup. A velocity dampening scalar of 0.08 was chosen because of drone limitations. Since the drones do not accept acceleration, velocity dampening reduces the likely hood of a drone moving as it approaches the target and velocity direction is not very reliable since the drones have built-in stabilization feature that can result in a reversed-velocity detection.

A total of 13 experiments were performed with various initial take off positions of the drones on the motion platform. Table 2 summarizes all the test cases and results of each experiment.

The initial drone positions are referred to as scattered, R-bar or V-bar based on the placement of them on the motion platform. The initial attitude of the RSO is written in °/s and the RSO maintained that initial spin throughout the duration of the respective test case. Status of each of the three drones at the end of the experiment are documented under the Drone 1, Drone 2 and Drone 3 columns. Finally, if any of the drones failed to dock, the reason for failure is mentioned in the failure reasoning column and the failure is discussed in detail in the Chaser Mission Failure sub-section.

**Table 2: Experimental Test Results**

| Test Name | Drone Positions | Yaw [°/s] | Pitch [°/s] | Roll [°/s] | Drone 1 | Drone 2 | Drone 3 | Failure Reasoning |
|---|---|---|---|---|---|---|---|---|
| Test 1 | Scattered | 0 | 0 | 0 | Docked | Docked | Docked | - |
| Test 2 | R-bar and V-bar | 0 | 0 | 0 | Docked | Docked | Inspection Orbit | - |
| Test 3 | Scattered | 0 | 0 | 0 | Docked | Docked | Failed | IMU failed |
| Test 4 | Scattered | 0 | 0 | 0 | Docked | Docked | Docked | - |
| Test 5 | Extreme positions and Scattered | 0 | 0 | 0 | Docked | Failed | Docked | RealSense error |
| Test 6 | Scattered | 1 | 0 | 0 | Docked | Docked | Docked | - |
| Test 7 | Scattered | 1 | 0 | 0 | Docked | Docked | Failed | IMU Failed |
| Test 8 | R-bar | 1 | 0 | 0 | Docked | Docked | Docked | - |
| Test 9 | V-bar | 1 | 0 | 0 | Inspection Orbit | Docked | Docked | - |
| Test 10 | Extreme positions and Scattered | 1 | 0 | 0 | Docked | Docked | Docked | - |
| Test 11 | Scattered | 5 | 0 | 0 | Inspection Orbit | Docked | Failed | OptiTrack Error |
| Test 12 | Scattered | 5 | 0 | 0 | Docked | Docked | Docked | - |
| Test 13 | R-bar | 5 | 0 | 0 | Docked | Docked | Docked | - |

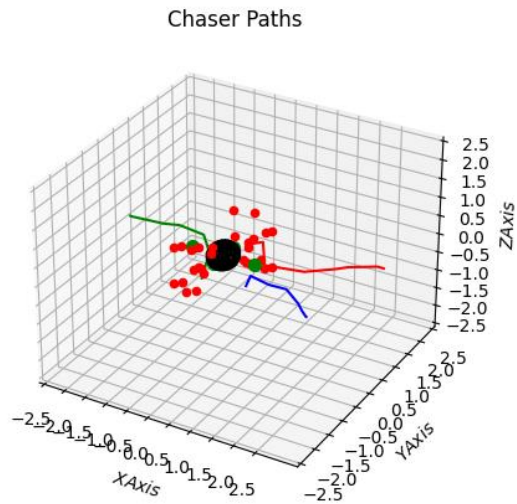

**Figure 9: Test 7 Chaser Graph**

**Docked**

A "Docked" condition was satisfied when a chaser is within a predetermined range of the two primary attractive nodes for a predetermined amount of time. For the first thirteen tests, the range was half a meter. This value was chosen due to half the range being within the physical target spacecraft and the visual limitations of OptiTrack: when chasers get too close to the target, the chaser is no longer visible to more cameras, increasing error in the live position. The primary attractive nodes are the centroids of the target spacecraft body, identified by the MV camera; in a real case, these faces would be desirable for chasers to either weld themselves or grab onto. The time required to dock is based on the number of cycles of the APF algorithm. The algorithm has a set delay of 0.25 seconds per cycle and the docking conditions were set to two cycles. Future iterations of the algorithm now have the docking range down to 0.3 meters and the cycles to dock at ten cycles. Additionally, the two docking ports on the other faces of the target body will be added. The additional docking ports represent the arm extenders for the solar panels, a rigid structure that chasers can grab onto.

**Inspection Orbit**

In the orbital scenario, the inspection orbit would occur after all the available docking ports are occupied or if the chaser is locked in the equivalent of a Lagrange points in the APF. The reason these mock Lagrange points can exist, is due to the limitations of movement; if the APF produces a resultant acceleration smaller than what the chaser is able to produce, the chaser will remain stuck if nothing is done; Tello Drones are limited to movement greater than 20 cm. Inspection orbits primarily only occur with the last chaser, since Chaser-Chaser repulsiveness typically is enough to push an idle chaser into action again. However, an inspection orbit is not a disadvantageous event. In orbital scenarios, the chasers would be outfitted with the computer vision technology and predetermined chasers would act as cameras maintaining relative positions to one another. If an approaching chaser was set into an inspection orbit, it would become an additional camera to strengthen the reliability of the target image. Since the algorithm has been updated, it is now able to recognize inspection orbits on its own and doubles the time differential used for calculating the acceleration to boost the position change.

**Discussion**

The research had a 70% success rate out of 13 tests. Though the failure rate seems high, the failed test cases were primarily a result of hardware failure and not implementation failure. These failures could have been mitigated with better and more reliable hardware. The experiments showed that machine vision and APF subsystems can collaboratively perform together to communicate with the drones to successfully navigate to potential docking site on the RSO, both tumbling and steady. Throughout the experiments, at least two of the chasers successfully docked with the target or remained in an inspection orbit.

**Table 3: Chaser Mission Failure**

| Reason | Explanation |
|---|---|
| IMU Failed | Hardware limitation/failure. If the lighting was not right, the infrared camera on the bottom of the drones would stop stabilizing the drone and the drones would drift after movements, often resulting in a crash. Drone 3 also had a history of bad crashes during the early stages of testing that lead to some damage to the IMU as well. |
| OptiTrack Error | OptiTrack relies on small reflective nodes to identify rigid bodies, the reflective foil on the target satellite can also register as nodes. When targets go out of view of the cameras, sometimes the |

| | | |
|---|---|---|
| | | motive software misinterprets the foil as the chaser and the APF algorithm adjust accordingly. |
| | RealSense Error | It was noticed that the RealSense camera caused the drones to collide due to its uncertainty in depth. This issue can be sorted out if the chasers/drones are also equipped with an additional depth sensor or vision-based navigation while they continue to rely on the mothership (in this case OptiTrack) for guidance. |

## LIMITATIONS AND FUTURE WORK

Though the current implementation works for certain test cases such as – RSO is yawing only, if the RSO is yawing and pitching the implementation will fail. These limitations are a result of current hardware or software limitations that are currently being accounted for and improved upon.

### Single Camera View

One of the greatest current limitations is the single view of the target. The APF algorithm is built to reflect the single input 180 degrees to produce a two-camera view, but ultimately only one input is feeding the algorithm. As a result, any uncertainty in machine vision is directly translated into APF. To address the issue, three cameras will be implemented at different angles to produce a more confident image of the target and to avoid uncertainty in one camera from impacting the confidence of APF.

### Stationary Docking Port

As a result of single camera view, a consistent image of the target body is not reliable. As a result, the algorithm currently sets the attractive nodes/ docking ports on the x-axis based on the target dimensions. As the target rotates, the chasers cannot approach when the repulsive nodes are in line with the docking port and wait until the repulsive nodes have moved away before making their final approach. As a result, however, tumbling tests only work with yawing motion. Once more cameras are added to the system, a more confident image of the target will allow for moving docking ports and more docking ports. With a docking port moving in all directions, both rolling and pitching motion will be possible.

### Drone Overshoot

Since the drones are not equipped with a robust control system the drones tend to overshoot to get to a specific location in the trajectory causing instability in the trajectories leading to an oscillation motion and even collision with the target satellite. This limitation can be resolved with a robust control system.

### Lighting / Drone IMU Issue

Always need overhead lights because the drones use IR sensors to calibrate its IMU. However, since the lab is all black to simulate space like environment, if there is not enough lighting the IR sensor fails to work resulting in failure of the IMU sensor. This issue required all the tests to be performed with overhead lighting. In contrast, the current machine vision algorithm works to its potential under darker lighting conditions as studied in [9, 28, 30]. This resulted in missed detections in some to most of the rotating spacecraft tests.

### Target Size and Scale

The target satellite has attractive nodes very close, not even one drone length away from the solar panel. During earlier tests, this caused the drones trying to dock with that attractive node to collide with the solar panels. That is part of the reason those nodes were not considered docking ports for early iterations of MARVIN. The issue was not entirely due to sizing, but also uncertainty in machine vision when rotating the target.

However, in a real-life scenario this will not be an issue if the chasers are smaller than the minimum distance between the nodes. Additionally, if the chasers are also equipped with depth detection sensors along with vision system, chaser-target collision can be preventable every time

**Machine Vision Framerate**

With the current hardware setup, the machine vision algorithm runs at a speed of 2FPS. In real-life docking missions, 2FPS is very slow especially when the spacecraft are so close to each other by a tumbling RSO. Such slow speeds could lead to catastrophic collisions. These slow processing speeds are due to machine vision's non-max suppression running on the CPU instead of the NCS2 due to hardware drawback.

To battle this drawback, our future research focuses on generating timeseries of machine vision detections to predict the dynamics of the RSO. Additionally, with the upcoming advancements to spacecraft hardware [34], it would be reasonable to deploy the algorithms on more powerful processors than Raspberry Pi 4 such NVDIA Jetson Nano or Orin. Doing so would enhance the speed and performance of the algorithms since larger and more reliable neural networks can be trained and deployed to perform detection.

**CONCLUSION**

From the hardware-in-the-loop testing conducted at the ORION facility in Florida Institute of Technology, MARVIN proved that APF and machine vision can collaborate together to enable a swarm of three chaser drones to autonomously execute a safe trajectory to a non-cooperative yawing RSO. Though MARVIN has limitations, it was identified that the overall performance of can be improved by upgrading to better and more reliable hardware, adding timeseries to predict the dynamics of the non-cooperative RSO and adding multiple camera feeds to cover the entirely of the non-cooperative RSO for detections such that future experiments are not limited to the RSO only yawing. Overall, proving this concept enables a novel approach to solving numerous problems ranging from OOS, ADR to preserving national security.

**ACKNOWLEDGMENTS**

This project was supported by the AFWERX STTR Phase II contract FA864921P1506 and the NVIDIA Applied Research Accelerator Program.